\documentclass[runningheads]{llncs}
\usepackage{graphicx}
\usepackage{caption}
\usepackage{amsmath,amssymb} 
\usepackage{color}

\usepackage{float} 
\usepackage{graphicx}
\usepackage{amsmath}
\usepackage{comment}
\usepackage{booktabs}
\usepackage{enumitem}
\usepackage{makecell,multirow,tabularx}
\usepackage{booktabs}
\usepackage{multirow}
\usepackage{colortbl}
\usepackage{times}
\usepackage{multirow}
\usepackage{algorithm}
\usepackage{algpseudocode}
 \usepackage[table,xcdraw]{xcolor}

\usepackage{epsfig}
\usepackage{subcaption}
\usepackage[pagebackref=true,breaklinks=true,letterpaper=true,colorlinks,bookmarks=false]{hyperref}

\begin{document}
\pagestyle{headings}
\mainmatter
\def\ECCV16SubNumber{***}  

\title{Empowering Morphing Attack Detection using Interpretable Image-Text Foundation Model} 




\author{Sushrut  Patwardhan \inst{1,2} \and Raghavendra Ramachandra \inst{1}\orcidID{0000-0003-0484-3956} \and
Sushma Venkatesh \inst{2}\orcidID{0000--0002-8557-0314} 
}
\authorrunning{Sushrut, Raghavendra \& Sushma}
%
\institute{Norwegian University of Science and Technology (NTNU), Norway. \\ \and
MOBAI AS, Norway. \\
\email{Email: raghavendra.ramachandra @ntnu.no}\\
}
\maketitle

\begin{abstract}
Morphing attack detection has become an essential component of face recognition systems for ensuring a reliable verification scenario.  In this paper, we present a multimodal learning approach that can provide a textual description of morphing attack detection. We first show that zero-shot evaluation of the proposed framework using Contrastive Language-Image Pretraining (CLIP) can yield not only generalizable morphing attack detection, but also predict the most relevant text snippet. We  present an extensive analysis of ten different textual prompts that include both short and long textual prompts. These prompts are engineered by considering the human understandable textual snippet. Extensive experiments were performed on a face morphing dataset that was developed using a publicly available face biometric dataset. We present an evaluation of SOTA pre-trained neural networks together with the proposed framework in the zero-shot evaluation of five different morphing generation techniques that are captured in three different mediums. 
\end{abstract}

\section{Introduction}
\label{sec:intro}
Face Recognition Systems (FRS) are widely used in high-security applications such as border control for seamless and accurate verification. However, they are vulnerable to various attacks, with morphing attacks posing a significant threat, particularly in real-life border control scenarios \cite{Venkatesh-MADSurvey-IEEETTS-2021}, \cite{ramachandra2024multispectral}, \cite{singh20233d}, \cite{10162056}, \cite{10157461}. Morphing involves blending the facial images of multiple subjects to create a new image that retains the characteristics of each, making it difficult for both human operators and automated FRS to distinguish them from legitimate faces, as demonstrated in \cite{godage2022analyzing}. This poses a serious risk in identity-document scenarios, where morphed images can compromise the integrity of the document by linking it to multiple individuals, potentially leading to misuse. Therefore, morphed image detection is essential to ensure reliable user verification.

Face-morphing attack detection (MAD) has been extensively studied in the literature and can be broadly classified as (a) single image-based MAD (S-MAD) and (b) differential-based MAD (D-MAD).  The S-MAD approach uses a single image, whereas D-MAD uses two images (one from a passport and another from a trusted device) to determine the morphing. Among these two approaches, the S-MAD-based approach reports a higher error rate than the D-MAD approaches \cite{Nist-Frvt-Morph}.  In our work, we focus exclusively on S-MAD approaches, as these methods present unique challenges in terms of interpretation and detection. Although deep learning-based MAD algorithms have demonstrated reasonable detection accuracy, their lack of interpretation limits their reliability. Consequently, extensive research has been conducted on S-MAD-based interpretation techniques \cite{10157534}.

Early work on interpretable MAD was focused on visualising the handcrafted features that can Early works on interpretable MAD focused on visualizing handcrafted features that can indicate the characteristic features of bona fide and morph images. The first work on interpreting handcrafted features is presented in \cite{Raghavendra-DetectingMorphedFace-BTAS-2016} which illustrates the features from the Binary Statistical Image Feature (BSIF) method for S-MAD. The interpretation of five different handcrafted features is presented in \cite{10157534} for S-MAD, which can indicate the different patterns of  texture corresponding to bona fide and morph. The use of color channels as an interpretation technique for S-MAD was discussed in \cite{Raghavendra-DetectingFaceMorphing-CVIP-2018} which shows the relevance of color channels. However, the use of handcrafted features as the interpretation does not provide useful information for non-technical personals, such as border guards, to determine the reason behind the decision of the S-MAD technique. The first work on the interpretation of deep-learning-based approaches for S-MAD was presented in \cite{Raghavendra-DNNMorphingDetection-CVPR-2017}, \cite{10224168}. The gradient maps were used to interpret the deep features from different layers of the CNN, which indicated the unique features of the morph and bona fide images. In \cite{Venkatesh-MADSurvey-IEEETTS-2021}, the interpretation of residual noise-based S-MAD methods was presented for both deep learning and handcrafted methods. The interpretation of the residual deep network was presented in \cite{RAJA2022104535}, where heat maps were used to present the interpretation of S-MAD. The pixel based interpretation using Focused Layer-wise Relevance Propagation (FLRP) are proposed in \cite{computers10090117}. The VGG-A based DNN architecture is used to train the morphing attack detector to evaluate the FLRP to indicate the pixel level explainability on the decisions.

Based on the aforementioned discussion, heat maps are commonly used as an explanatory tool for decision-making processes. However, it is important to note that heat maps generated using GradCAM \cite{GradCam} or similar methods may only highlight specific areas within an image and can exhibit a noisy nature. As a result, individuals without technical expertise, particularly those working in border control agencies, may find it challenging to interpret these approaches. Therefore, the visual explanations have several disadvantages, they are not precise, fine-grained, and may be subject to interpretation. 

In this work, we propose a multimodal approach to explain the decision of the S-MAD. To the best of our knowledge, this is the first work to provide a textual description of the decision to morphing attack using approaches based on deep learning. We propose the use of multimodal approach Contrastive Language-Image Pre-training (CLIP) \cite{radford2021learning} that can predict the most relevant text snippet for a given morphing or bona fide face image. We present zero-shot learning on CLIP with different text prompts to obtain human understanding-level explanations.  The following are the main contributions of this work: 

\begin{itemize}  [leftmargin=*,noitemsep, topsep=0pt,parsep=0pt,partopsep=0pt]
\item We present the textual explainable method for face morphing attack detection using zero-shot learning of Contrastive Language-Image Pre-training (CLIP) \cite{radford2021learning} especially with zero shot settings. To our knowledge, this is the first work that will predict the text snippet and the corresponding score for the morphing detection. 
\item We analyse ten different prompts balancing both coherent textual explanation and the morphing attack detection performance. This provide the basis on identifying the suitable prompt in zero shot settings.
\item Extensive experiments are presented with five different types of morphing generation techniques and three different types of mediums (digital, high quality print scan and low quality print scan). 
\end{itemize}

The rest of the paper is organized as follows: Section \ref{sec:Pro} presents the proposed method for explainable morphing attack detection, Section \ref{sec:exp} presents the qualitative and quantitative results with ten different prompts, and Section \ref{sec:conc} draws the conclusion. 
\section{Proposed Approach}
\label{sec:Pro}
The goal of the proposed approach is to enable the textual description of a single image-based morphing attack detection. Therefore, we are motivated to employ a multimodal framework based on Contrastive Language-Image Pre-training (CLIP) \cite{radford2021learning}. CLIP is an image-language model trained using a large number of image-text pairs based on a contrastive learning framework. One of the primary advantages of the CLIP model is its ability to perform well in zero-shot learning, which has led to its use in a variety of applications. In particular, recent studies have shown that CLIP can be used effectively for detecting facial presentation attacks \cite{srivatsan2023flip}. In \cite{srivatsan2023flip}, CLIP demonstrated a reasonable detection performance in cross-data scenarios and predicted textual snippet reflecting the decisions, particularly in the context of zero-shot learning. These results  motivated us to explore the use of CLIP in  zero-shot settings for single image-based MAD.
\begin{figure*}[htp]
  \centering
  \includegraphics[width=0.8\linewidth]{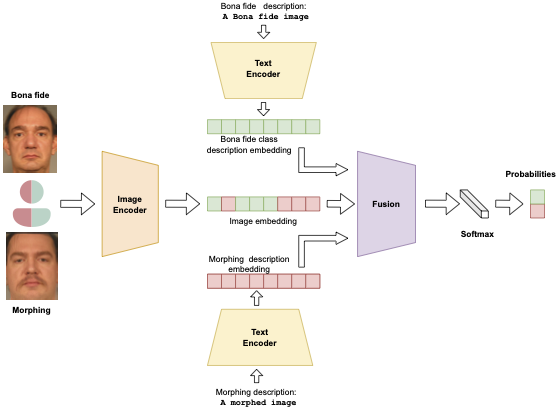}
  \caption{Block Diagram of the proposed framework using zero-shot explainable single image based morphing attack detection}
  \label{fig:prop}
\end{figure*}

Figure \ref{fig:prop} shows the block diagram of the proposed framework for textual explainability using multimodal models for single image based morphing attack detection.  The proposed framework is based on the  CLIP operating in the zero-shot settings to obtain both detection accuracy from the image encoder and textual explanation on decision using text encoder. We first presents the details of CLIP model followed by its utilisation for explainable morphing attack detection. 
\subsection{Contrastive Language-Image Pre-training (CLIP)}
The Contrastive Language-Image Pre-training (CLIP) neural network \cite{clip}  is trained using  millions of vivid images with readily available natural language supervision found on the Internet. Using this approach, the CLIP model effectively encodes both the image and its corresponding description in a shared feature space. The CLIP model comprises two primary components: an Image Encoder and the Text Encoder.
\subsubsection{Image Encoder}
The CLIP model \cite{clip} has two variations in its image encoder: ResNet-50 \cite{resnet50} and Vision Transformer \cite{vit}. Both variations have minor architectural adjustments, and for our method, we utilize a Vision Transformer-based image encoder, which has demonstrated superior performance, as detailed in \cite{clip}. Additionally, the use of pre-trained vision transformers has also shown improved generalization capability over other S-MAD techniques \cite{zhang2024generalized}. The image encoder consists of transformer blocks that generate image embeddings, which are then projected onto a shared feature space with text description using a linear layer.
\subsubsection{Text Encoder}
The text encoder implemented in CLIP features a transformer-based architecture with a few modifications, as discussed in \cite{transformer_text}. The encoder operates on a byte pair encoding (BPE) representation with all characters in the lowercase to ensure more accurate encoding. The maximum sequence length was limited to 76 to optimize performance. Similar to the image encoder, the embeddings generated by the text encoder were projected onto a shared feature space using a linear layer.
\subsection{Zero-Shot evaluation for explainable S-MAD}
Zero-shot learning (ZSL) is a machine learning approach where a model is able to recognize and classify objects  it has never encountered before. In order to perform single-image Morph attack Detection with Proposed framework using CLIP models, we pass the facial image through the image encoder and obtain the corresponding features. Simultaneously, we send the text description of the bona fide and morphed image through the text encoder and obtain the text features. The probabilities are then calculated using a softmax layer based on the cosine distance of the image and text features. As it is a zero-shot evaluation, the weights from the pre-trained network are employed without any alterations.

The key aspect of zero-shot evaluation lies in selecting the appropriate text prompt. The research results presented in \cite{clip} emphasizes the significance of prompt engineering in optimizing the performance of the CLIP model for zero-shot evaluation. To this end, we present ten diverse prompts in this work, taking into account both human comprehension and detection capabilities. These prompts were examined in terms of their suitability for morphing detection. Specific details regarding the different prompts and their respective detection performances are provided in Section \ref{sec:exp}. 


\section{Experiments and Results}
\label{sec:exp}
In this section, we present and discuss both the quantitative and qualitative outcomes of the proposed explainable framework for S-MAD, which utilizes CLIP with a zero-short evaluation. Initially, we will delve into the morphing dataset that was used in this research. We then discuss the various evaluation protocols that can be used to compare the zero-short evaluation across different types of prompts. Finally, we present and analyze the quantitative results for each prompt.

\subsection{Morphing Dataset}
\label{sec:db}

In this section, we introduce the Morphing Dataset (MD), which was sampled from the publicly available FRGC V2 dataset \cite{Phillips-OverviewFaceRecognitionGrandChallengeFRGC-CVPR-2005}. We selected 143 subjects with neutral expressions and postures, following the recommendations of \cite{zhang-MIPGAN-TBIOM-2021}. The image-pair selection corresponding to different subjects used for morphing is also based on the list from  \cite{zhang-MIPGAN-TBIOM-2021}. 
The morphing images are generated by employed five different face morphing techniques for generating face morphs: landmark-based (LMA-I) \cite{Landmark-face-morph}, landmark with post-processing (LMA-II) \cite{Ferrara-TextureBlendingAndShapeWarpingInFaceMorphing-IEEE-BIOSIG-2019}, MIPGAN-2 \cite{zhang-MIPGAN-TBIOM-2021}, MorDiff \cite{MoDiff}, and PIPE \cite{PIPE}.

The MD dataset is comprised of three different types of medium: digital and two types of Print-Scan (PS). The digital version encompasses conventional morphing images, whereas the PS morphing images are re-digitized versions of these digital images. The inclusion of the PS version was motivated to reflect the passport issuance scenario in which printed passport images were accepted. In this work, we created both high-quality and low-quality print-scan images that can reflect real-life scenarios. High-quality PS images were created by printing digital bona fide and morph images using DNP printers \cite{DNP-Printer} (we refer to this as PS-1), whereas low-quality PS images (we refer as PS-2) were created using a RICOH printer. In both cases, the printed images were scanned using an office scanner to achieve a resolution of 300dpi, following the recommendations of ICAO \cite{ICAO-9303-p1-2015}.

\begin{figure*}[]
  \centering
  \includegraphics[width=1\linewidth]{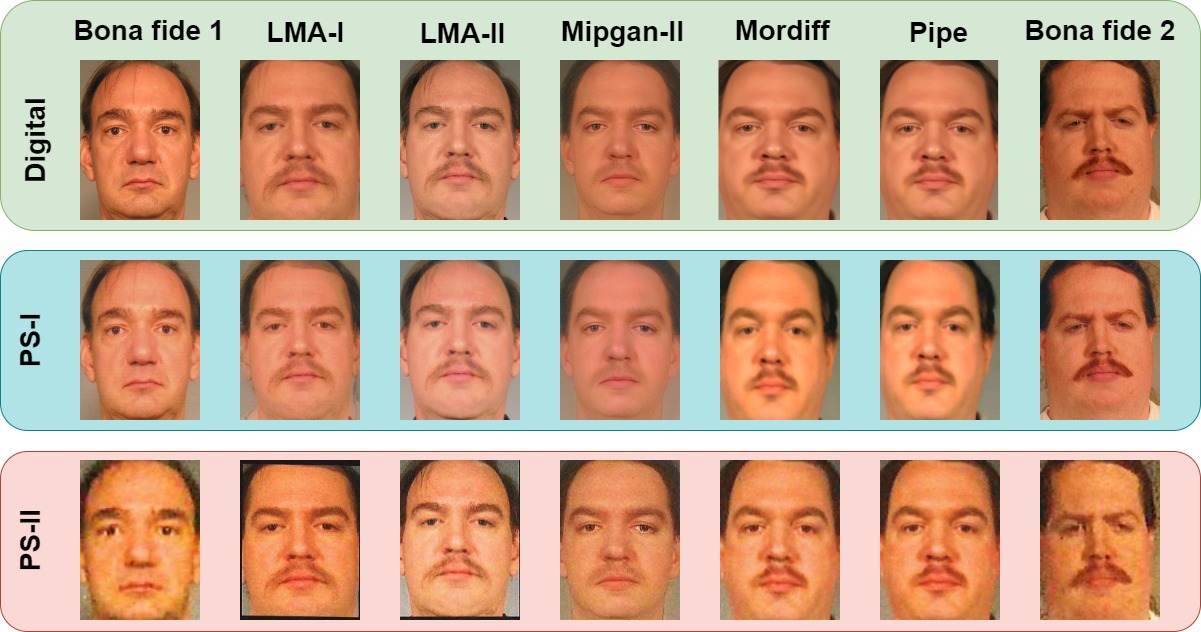}
  \caption{Illustration of morphing samples with different mediums from MD dataset.}
  \label{fig:DB}
\end{figure*}

Figures \ref{fig:DB} show examples of the MD dataset corresponding to the Digital and PS data sets. It is worth noting that the quality of the images was slightly degraded by PS. The MMD dataset contains 1276 bona fide samples (separately for digital and morphing) and 2526 morphing images (separately for five different morphing techniques, including the proposed method, and separately for digital and PS).  Therefore, the MD dataset has $1276 \times 2 = 2552$ bona fide samples and $2526 \times 5 \times 2 = 12630$ morphing images.

\subsection{Results and Discussion}
\label{sec:res}
\begin{figure*}[htp]
  \centering
  \includegraphics[width=1\linewidth]{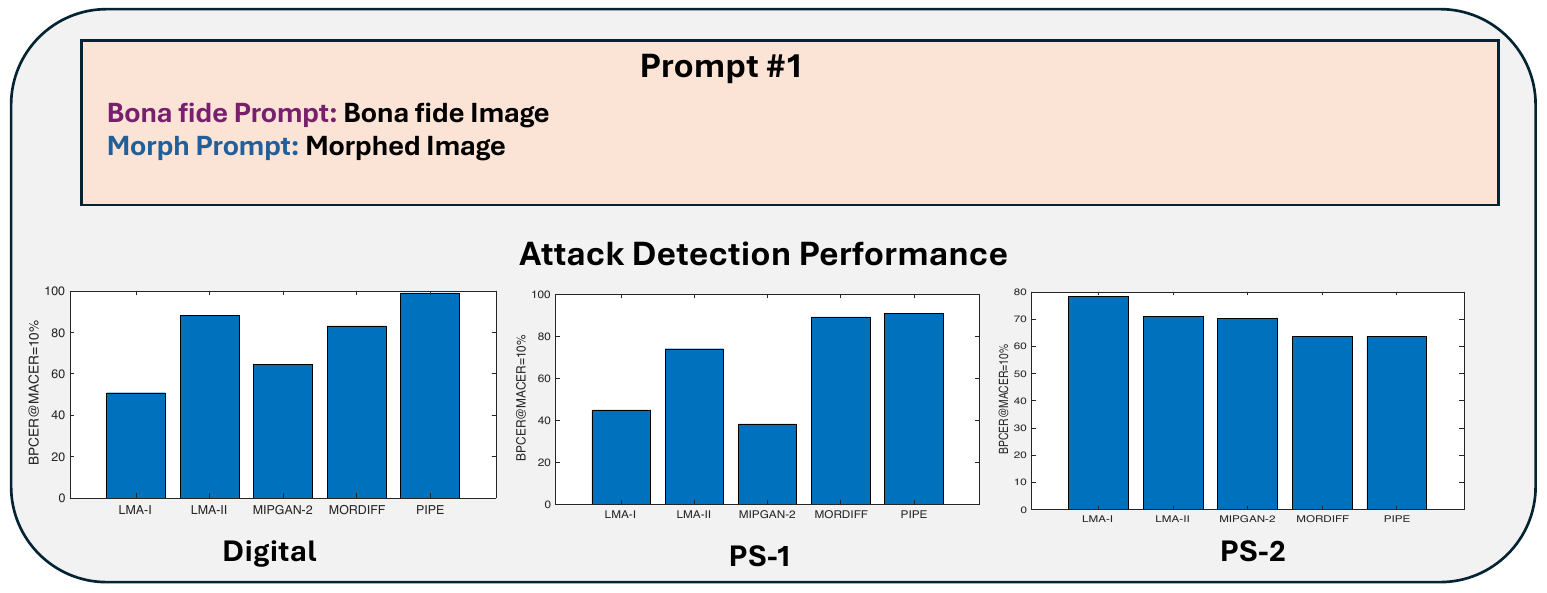}
  \caption{The quantitative results of the S-MAD  using Proposed Framework with Prompt \#1. }
  \label{fig:promp1}
\end{figure*}

\begin{figure*}[htp]
  \centering
  \includegraphics[width=\linewidth]{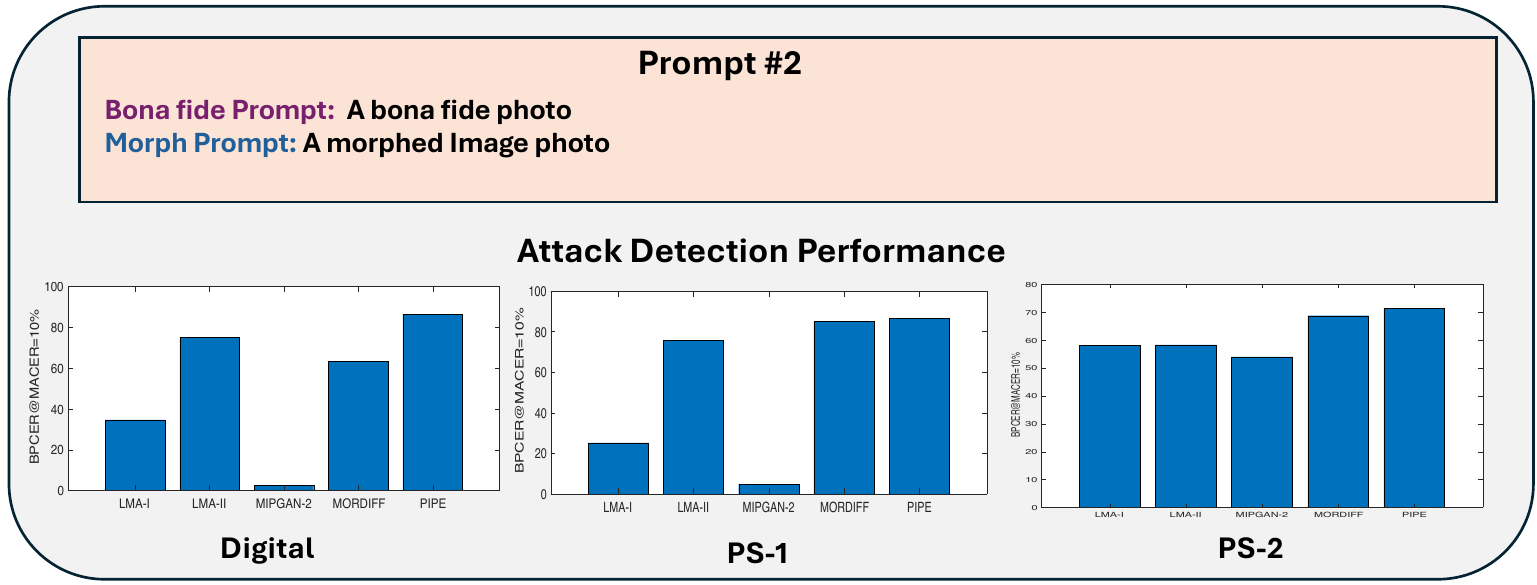}
  \caption{The quantitative results of the Proposed S-MAD Framework with Prompt \#2.}
  \label{fig:promp2}
\end{figure*}

\begin{figure*}[htp]
  \centering
  \includegraphics[width=\linewidth]{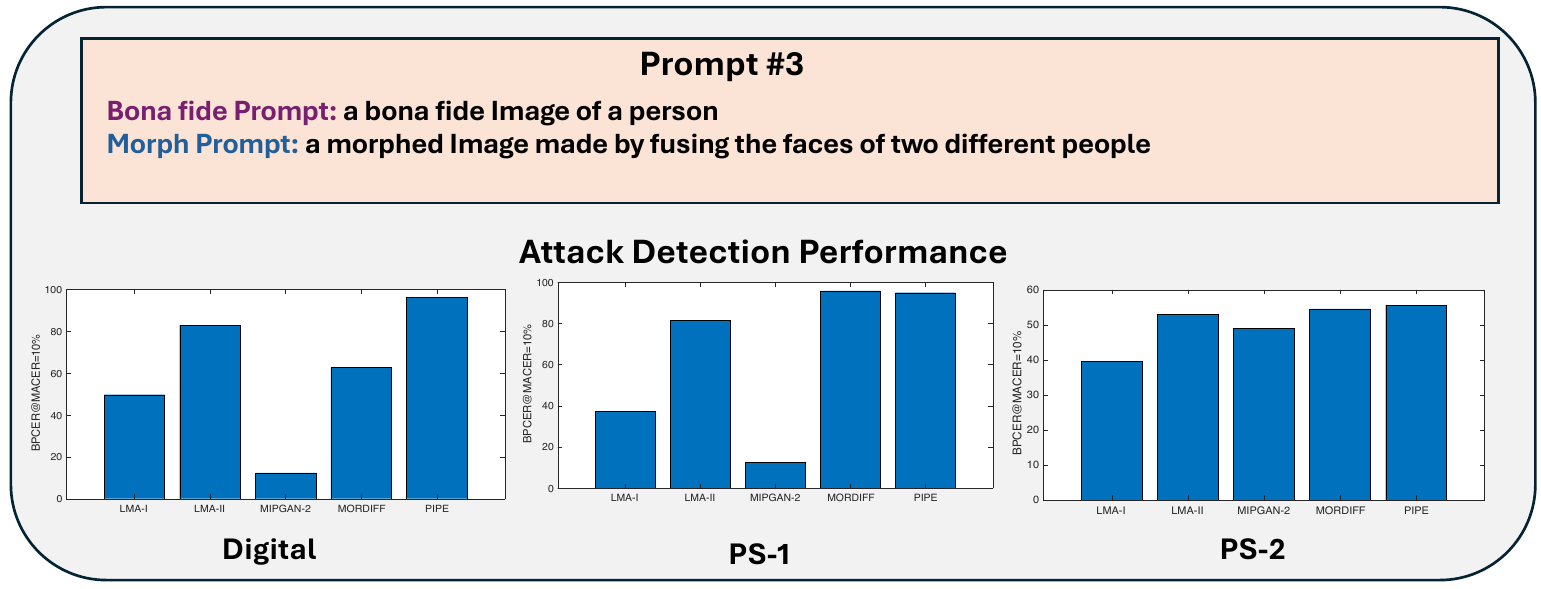}
  \caption{The quantitative results of the Proposed S-MAD Framework with Prompt \#3.}
  \label{fig:promp3}
\end{figure*}

\begin{figure*}[htp]
  \centering
  \includegraphics[width=\linewidth]{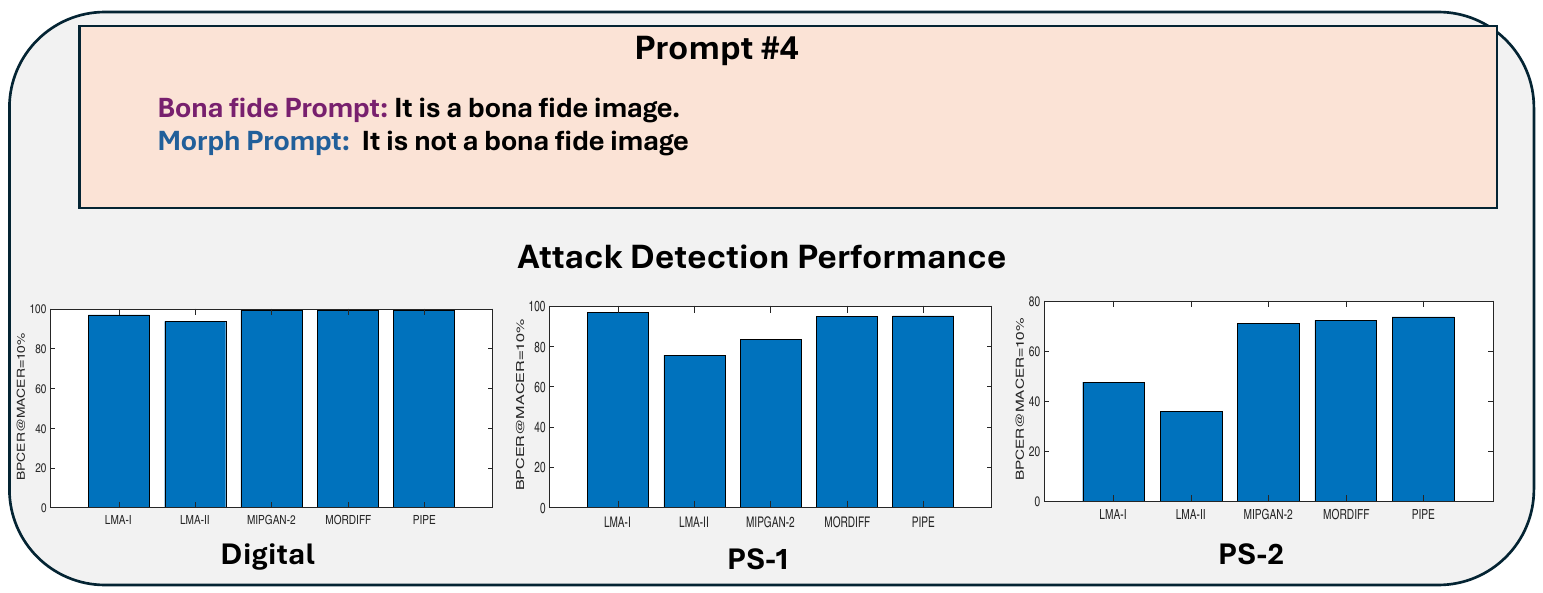}
  \caption{The quantitative results of the Proposed S-MAD Framework with Prompt \#4.}
  \label{fig:promp4}
\end{figure*}

\begin{figure*}[htp]
  \centering
  \includegraphics[width=\linewidth]{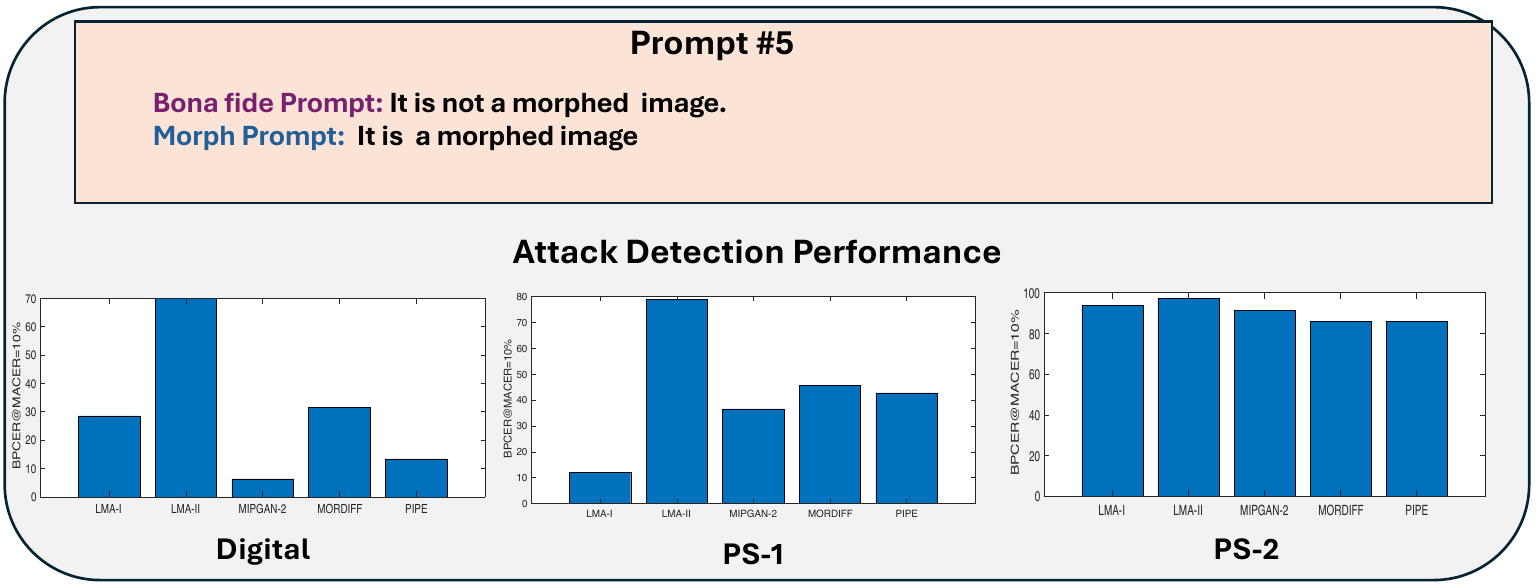}
  \caption{The quantitative results of the Proposed S-MAD Framework with Prompt \#5.}
  \label{fig:promp5}
\end{figure*}

\begin{figure*}[htp]
  \centering
  \includegraphics[width=\linewidth]{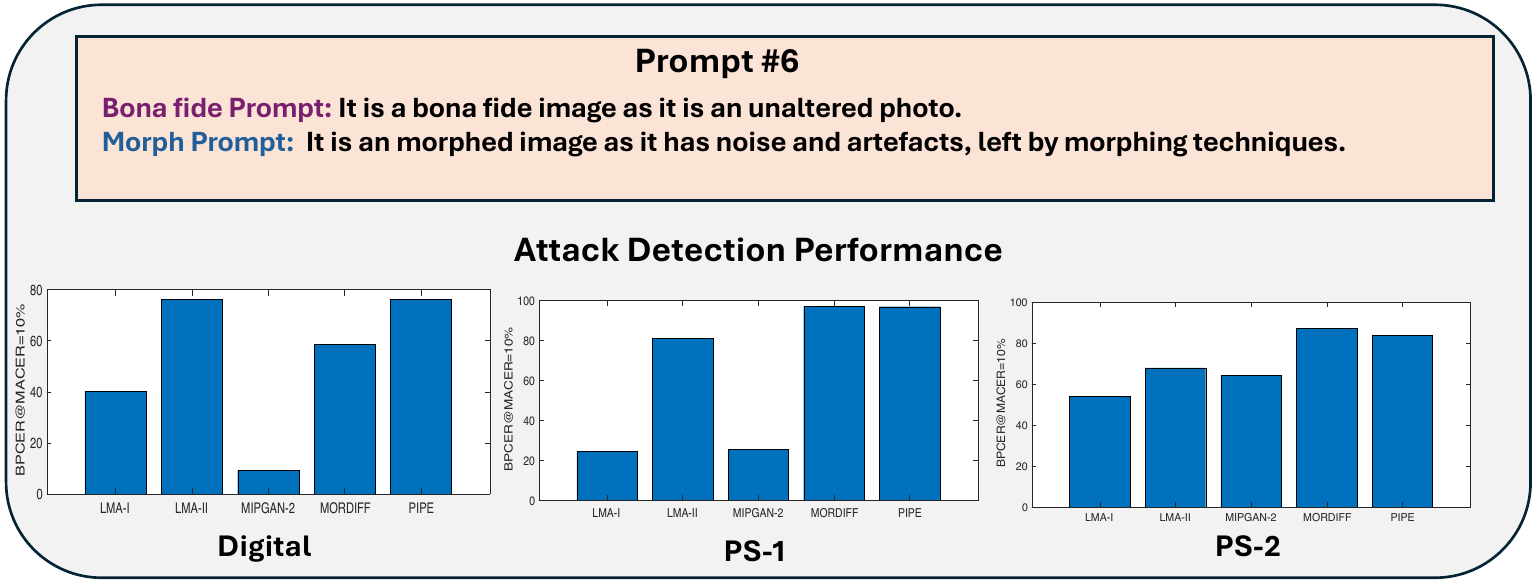}
  \caption{The quantitative results of the Proposed S-MAD Framework with Prompt \#6.}
  \label{fig:promp6}
\end{figure*}

\begin{figure*}[htp]
  \centering
  \includegraphics[width=\linewidth]{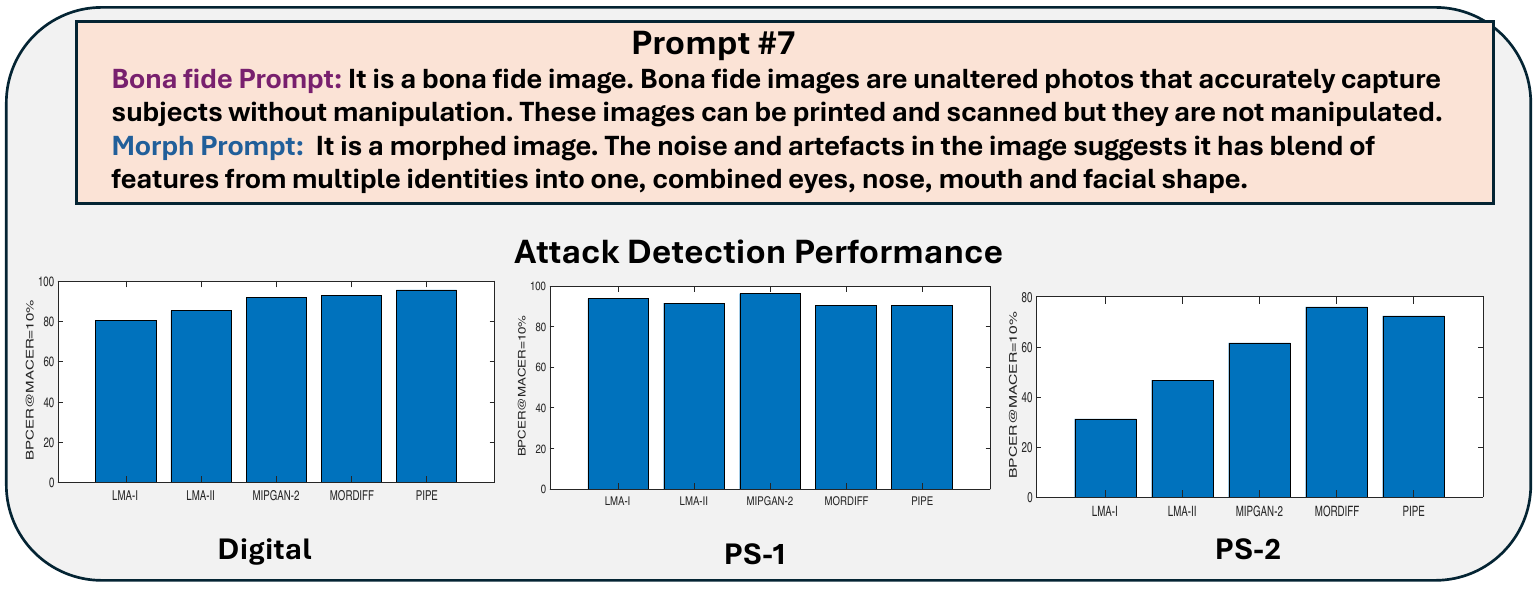}
  \caption{The quantitative results of the Proposed S-MAD Framework with Prompt \#7.}
  \label{fig:promp7}
\end{figure*}

\begin{figure*}[htp]
  \centering
  \includegraphics[width=\linewidth]{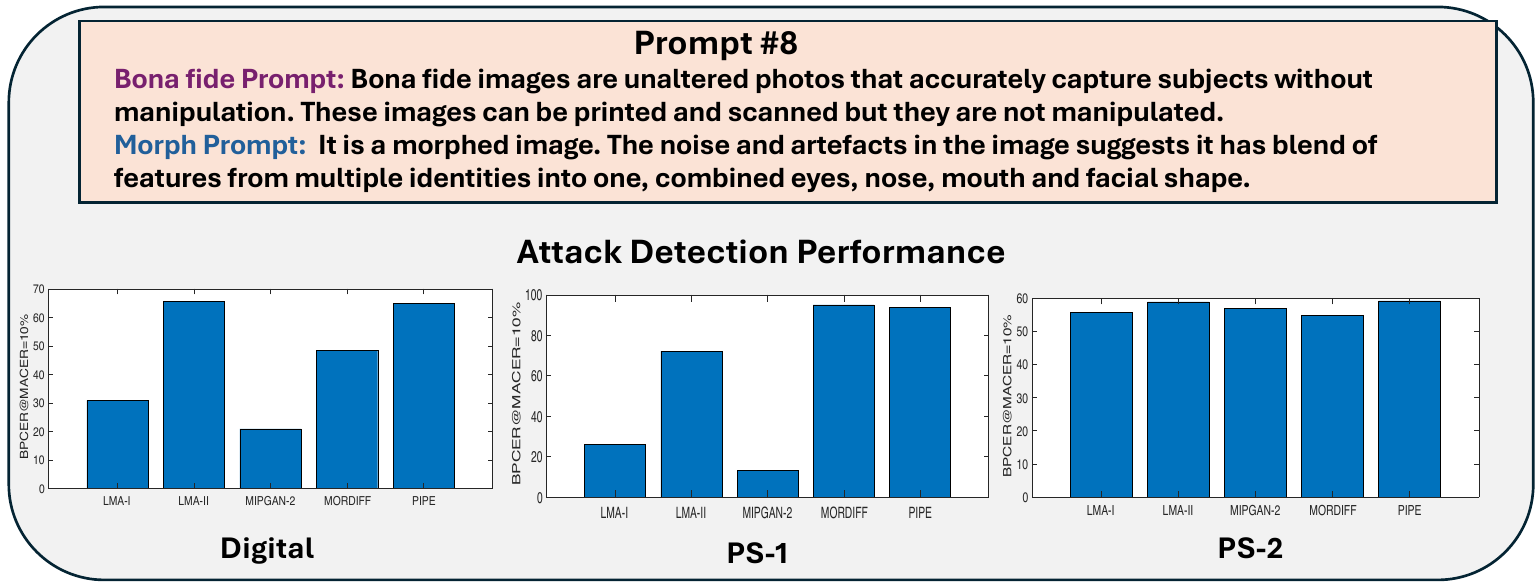}
  \caption{The quantitative results of the Proposed S-MAD Framework with Prompt \#8.}
  \label{fig:promp8}
\end{figure*}

\begin{figure*}[htp]
  \centering
  \includegraphics[width=\linewidth]{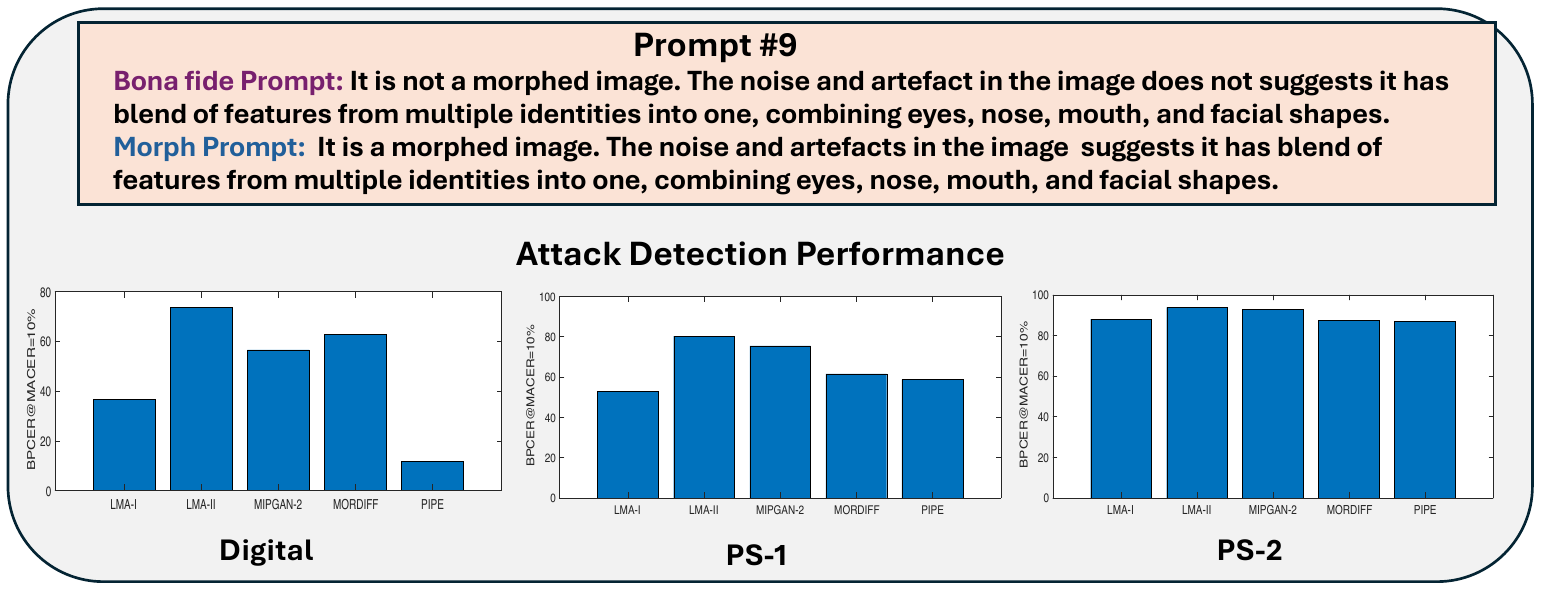}
  \caption{The quantitative results of the Proposed S-MAD Framework with Prompt \#9.}
  \label{fig:promp9}
\end{figure*}

\begin{figure*}[htp]
  \centering
  \includegraphics[width=\linewidth]{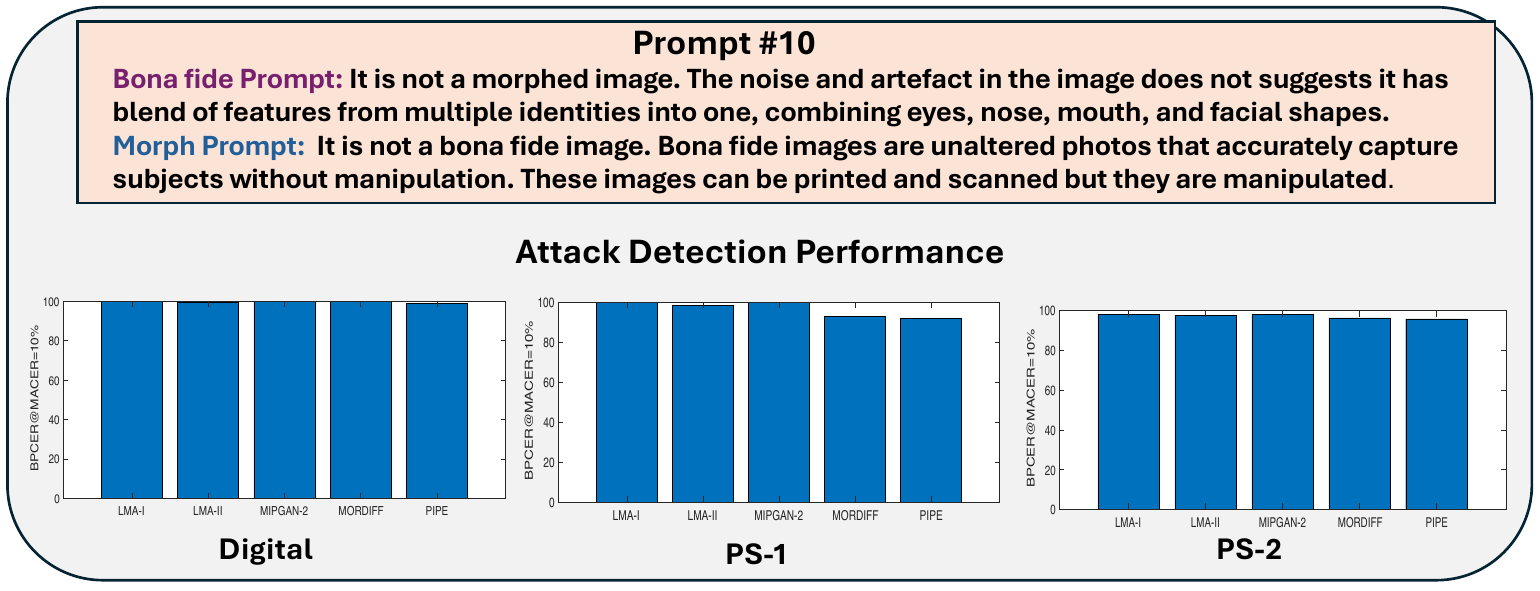}
  \caption{The quantitative results of the  Proposed S-MAD Framework with Prompt \#10.}
  \label{fig:promp10}
\end{figure*}

\begin{figure}[htp] 
    \centering
    \subfloat[Digital]{%
        \includegraphics[width=0.45\textwidth]{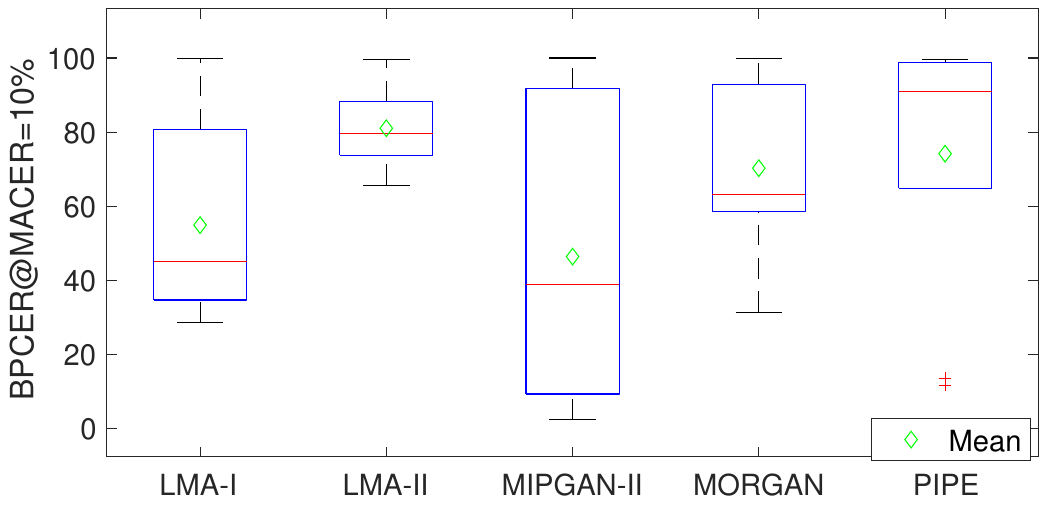}%
        \label{fig:E4aa}%
        }%
    \hfill%
    \subfloat[PS-1]{%
        \includegraphics[width=0.45\textwidth]{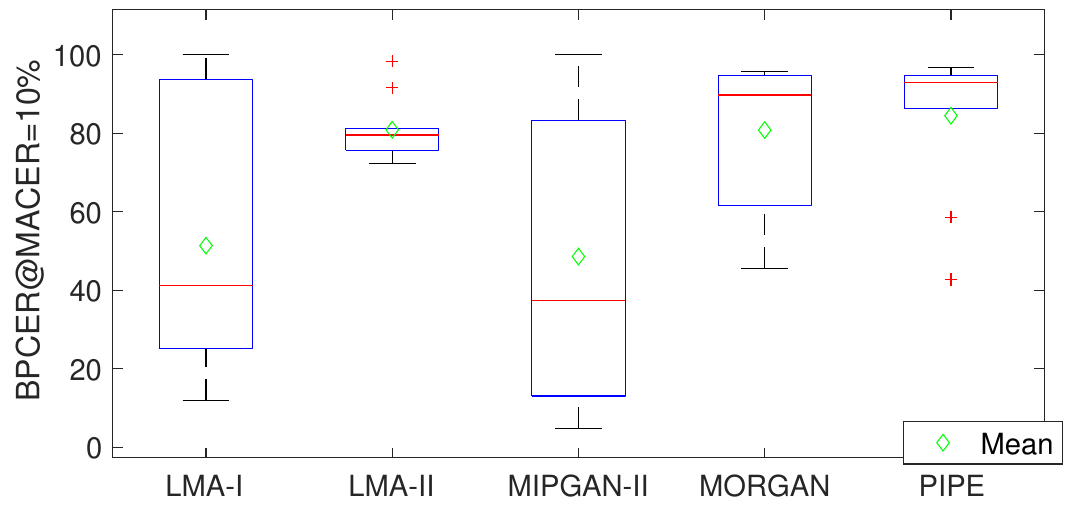}%
        \label{fig:E4bb}%
        }%
         \hfill%
    \subfloat[PS-2]{%
        \includegraphics[width=0.45\textwidth]{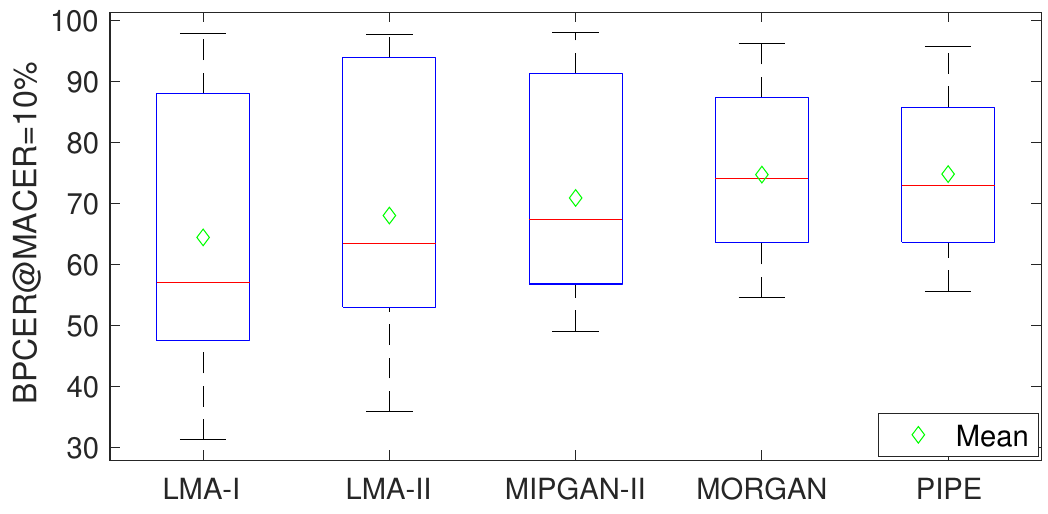}%
        \label{fig:E4cc}%
        }%
    \caption{Mean performance of different morphing generation algorithms across all prompts on each medium.}
     \label{fig:Exp4}
\end{figure}

\begin{figure}[htp] 
   \centering
    \subfloat[Shorter prompts]{%
       \includegraphics[width=0.45\textwidth]{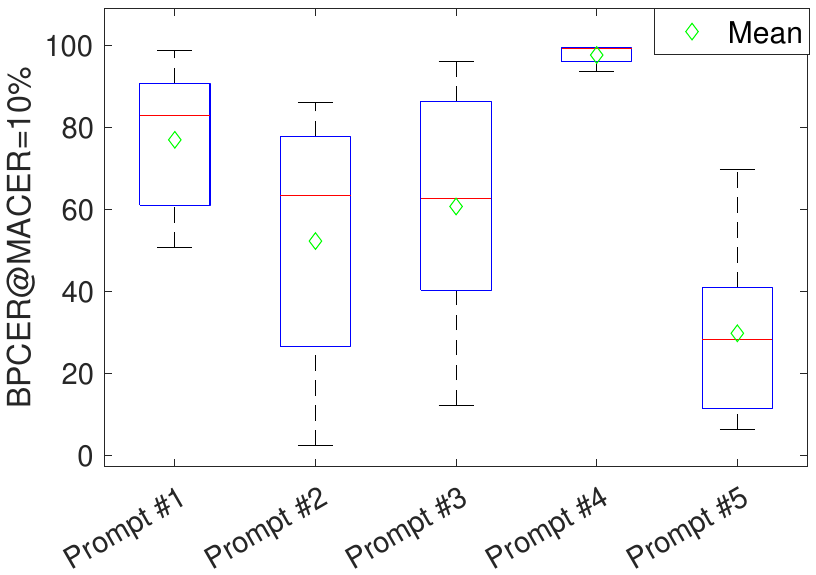}%
       \label{fig:a}%
       }%
   \hfill%
   \subfloat[Longer prompts]{%
       \includegraphics[width=0.45\textwidth]{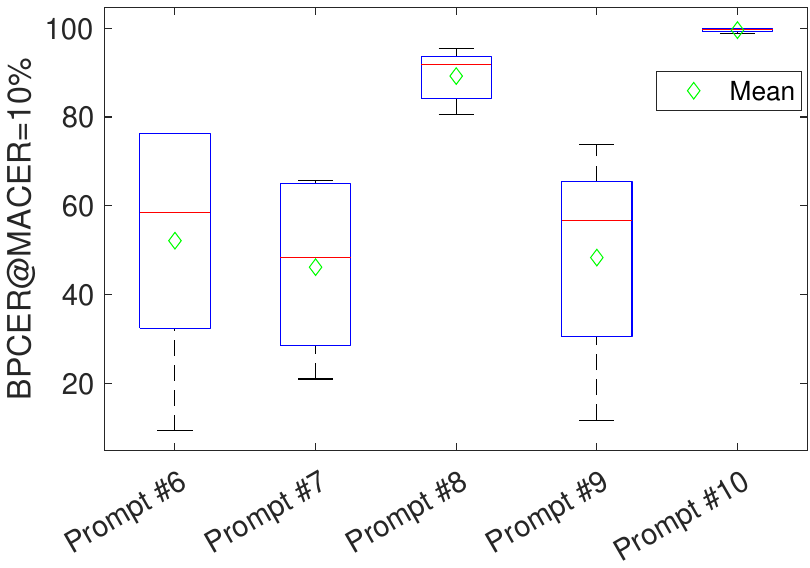}%
       \label{fig:b}%
       }%
   \caption{Mean performance of different morphing generation algorithms on shorter and longer prompts with digital medium.}
    \label{fig:Digi}
\end{figure}

\begin{figure}[htp] 
   \centering
   \subfloat[Shorter prompts]{%
       \includegraphics[width=0.45\textwidth]{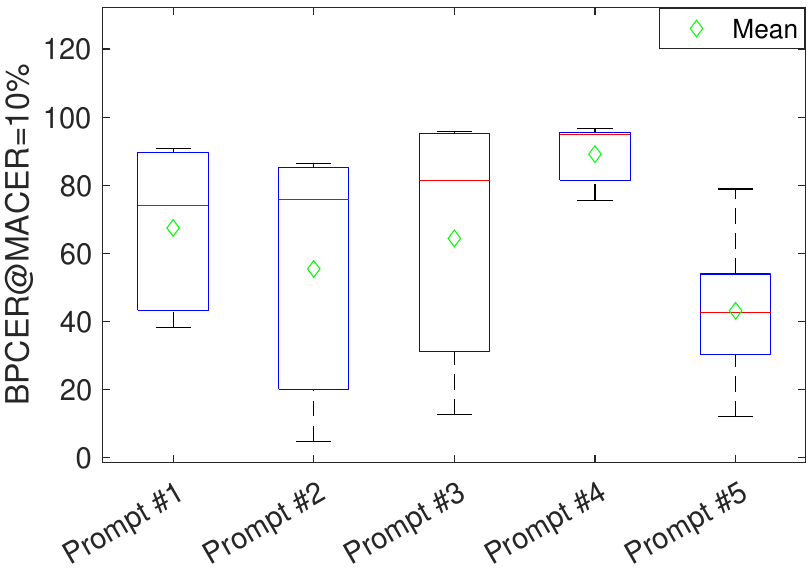}%
       \label{fig:c}%
       }%
   \hfill%
   \subfloat[Longer prompts]{%
       \includegraphics[width=0.45\textwidth]{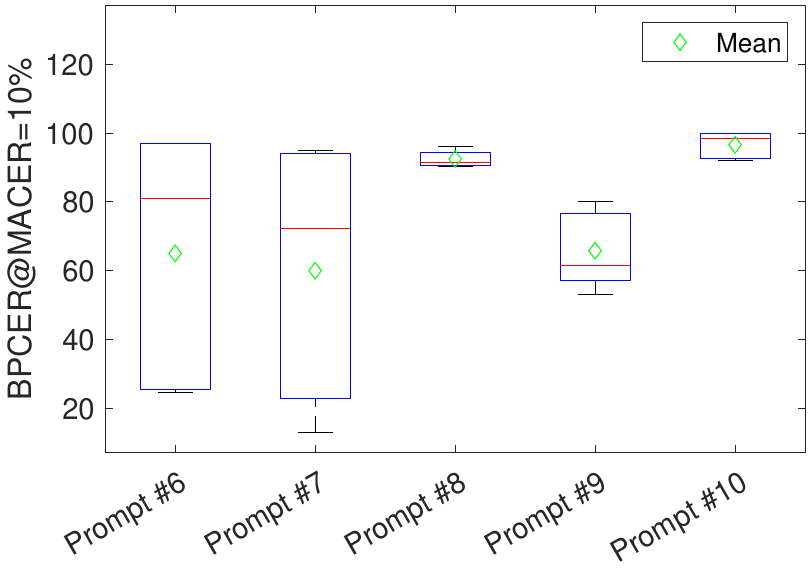}%
       \label{fig:d}%
       }%
   \caption{Mean performance of different morphing generation algorithms on shorter and longer prompts with PS-1 medium.}
    \label{fig:PS-I}
\end{figure}

\begin{figure}[htp] 
   \centering
   \subfloat[Shorter prompts]{%
       \includegraphics[width=0.45\textwidth]{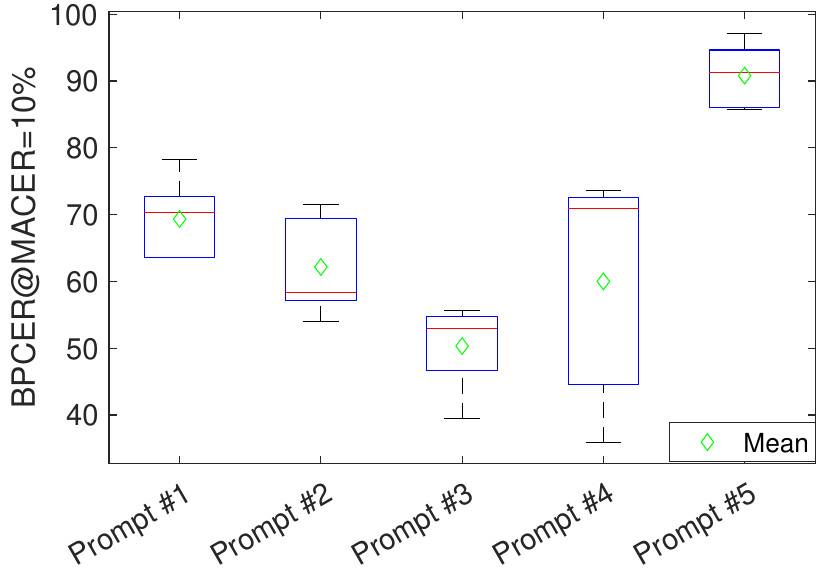}%
       \label{fig:e}%
       }%
   \hfill%
   \subfloat[Longer prompts]{%
       \includegraphics[width=0.45\textwidth]{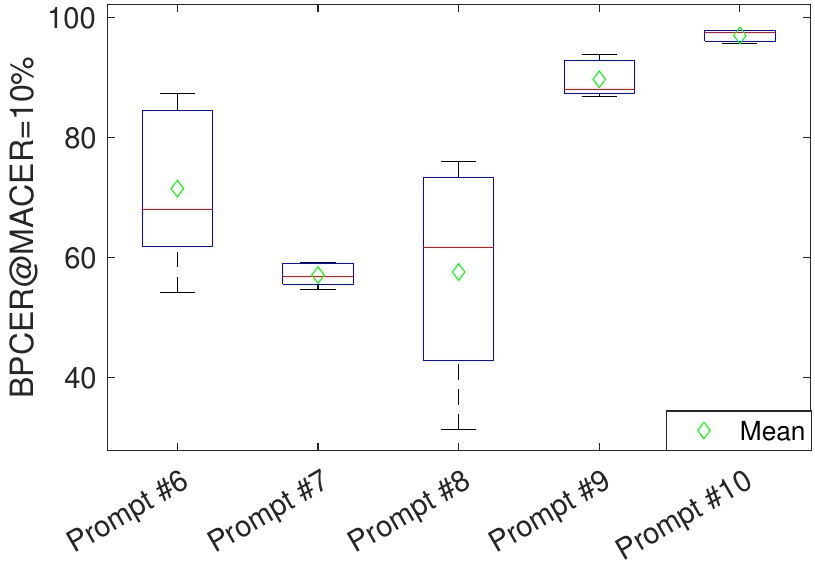}%
       \label{fig:f}%
       }%
   \caption{Mean performance of different morphing generation algorithms on shorter and longer prompts with PS-2 medium.}
    \label{fig:PS-II}
\end{figure}

\begin{figure*}[htp]
  \centering
  \includegraphics[width=0.85\linewidth]{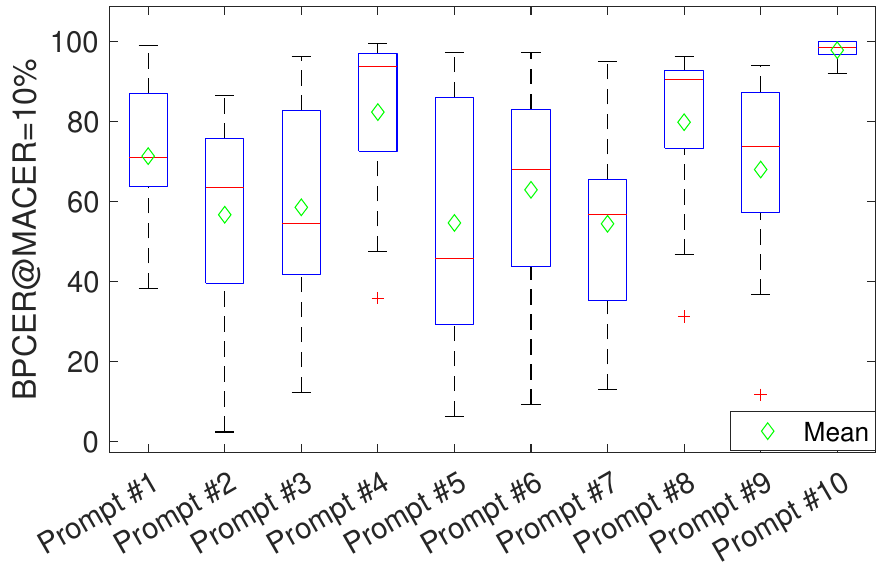}
  \caption{Mean performance of different morphing algorithms across all three mediums on individual prompts.}
  \label{fig:Exp2}
\end{figure*}

The quantitative results of the proposed framework are presented using ISO/IEC SC 37 30107 metrics, including the Morphing Attack Classification Error Rate (MACER) and the bona fide Presentation Classification Error Rate (BPCER). In this work, we present the results with BPCER @APCER = 10\%, corresponding to all ten prompts studied in this work for S-MAD. 
In this work, we present three different experimental evaluation protocols: (a) \textbf{Experiment-1:} This experiment aims to evaluate different text prompts by  balancing human understanding and detection accuracy.  Figures \ref{fig:promp1} – \ref{fig:promp10} discuss the quantitative accuracy of the prompts designed in this study. \textbf{Experiment-2:} This experiment aims to provide insights into quantifying the influence of detection based on prompt type (short or long), morphing generation algorithms, and morphing mediums.  Figure  \ref{fig:Exp4},  \ref{fig:Digi}, \ref{fig:PS-I}, and \ref{fig:PS-II} illustrate the quantitative results. \textbf{Experiment-3:} This experiment aimed to analyze the individual performance of the proposed prompts regardless of the generation and medium of the morphing process. Figure \ref{fig:Exp2} presents the quantitative results.

Figures \ref{fig:promp1} – \ref{fig:promp10} show the quantitative results corresponding to the individual prompts and their corresponding quantitative results for the three different data mediums. Among ten different prompts, the first five prompts (prompt \# 1–\#5) are engineered to have short sentences that can reflect similar (or synonyms) sentences that are used while training CLIP.  The last five prompts (Prompts \#6 - \#10) were engineered to make more descriptive and therefore these are long sentences. We present the detection results for five different morphing generation methods namely: land mark based approaches LMA-I \cite{Landmark-face-morph} and LMA-II \cite{Ferrara-TextureBlendingAndShapeWarpingInFaceMorphing-IEEE-BIOSIG-2019} and also generative methods such as MIPGAN-2 \cite{zhang-MIPGAN-TBIOM-2021}, Mordiff \cite{MoDiff} and PIPE \cite{PIPE}. The detection performance of the proposed framework for the S-MAD independently for each prompt is indicated in Figures \ref{fig:promp1} – \ref{fig:promp10}, in which both prompts that are used for each image using inference and the corresponding results are presented using bar charts. The x-axis in the bar chart represents the different morphing generation methods, and the y-axis indicates BPCER @MACER = 10\%. To provide better insights into the obtained results, we performed an additional analysis, as discussed below. 

Figure \ref{fig:Exp4} illustrates the average detection performance of individual morphing generation techniques across the three different mediums. Overall, the digital medium demonstrated superior detection performance compared to the PS-1 and PS-2 medium. Additionally, the performance varied among different morphing generation types. For the digital medium, MIPGAN-II and LMA-I achieved the best results, while a similar performance was observed for PS-1. However, for PS-2, LMA-I and LMA-II showed better performance than the generative methods used for morphing attacks. Therefore, the medium has an impact on the detection accuracy depending on the type of morphing-generation technique used.


\begin{figure}[htp] 
    \centering
    \subfloat[Shorter prompts]{%
        \includegraphics[width=0.45\textwidth]{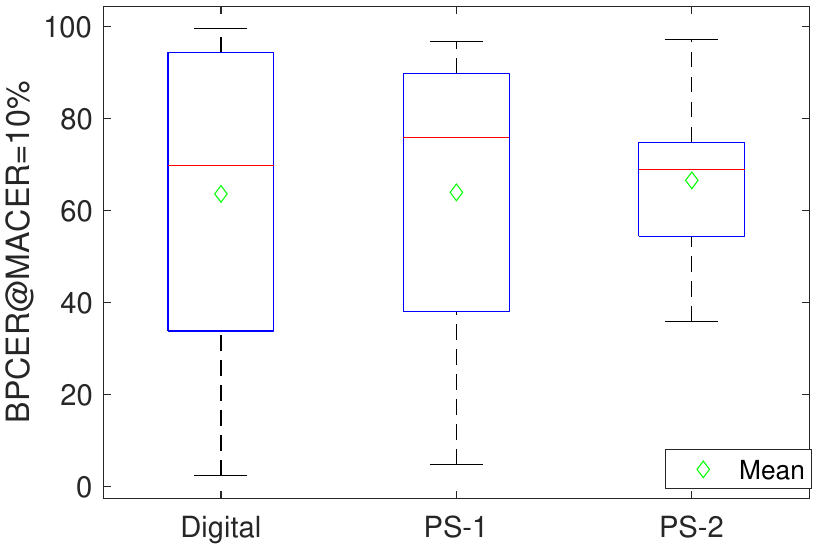}%
        \label{fig:aa}%
        }%
    \hfill%
    \subfloat[Longer prompts]{%
        \includegraphics[width=0.45\textwidth]{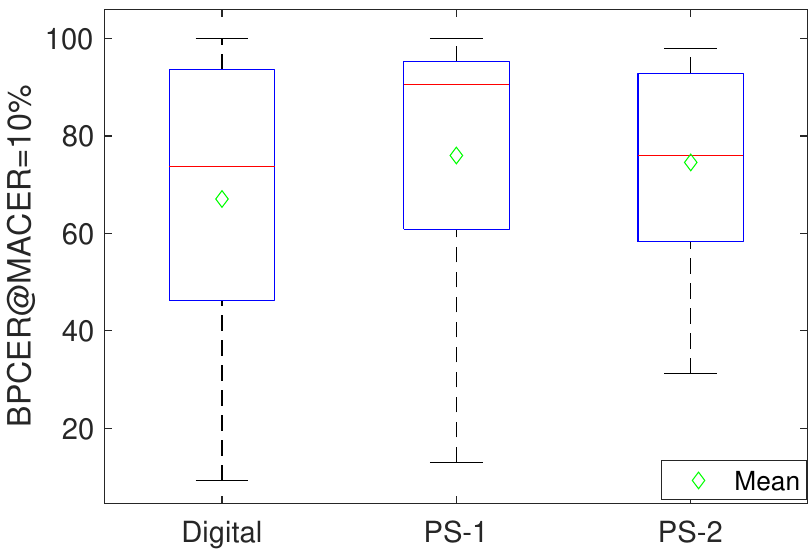}%
        \label{fig:bb}%
        }%
    \caption{Mean performance of different morphing algorithms across all prompts and morphing generation algorithms independently on each medium.}
     \label{fig:Exp3}
\end{figure}


Figure \ref{fig:Exp3} presents the individual box plots for each medium calculated by averaging the detection performance across both morphing generation algorithms and prompts. Overall, it is evident that shorter prompts generally lead to better detection performance than longer prompts. The digital and PS-1 mediums showed comparable performances for both short and long prompts. However, the PS-2 data medium exhibited similar performance for all three data media when using short prompts, but a decline in performance when using long prompts.

Figure \ref{fig:Exp2} shows the mean detection performance across both morphing  generation algorithms and  the medium with individual prompts. These results provide an overview of zero-shot evaluation in a real-life scenario where both the medium and generation types are unknown. The results indicated the best performance with prompt \#5, indicating the superiority of using a short prompt.

\begin{table}[htp]
\center
\resizebox{0.75\textwidth}{!}{
\begin{tabular}{|c|c|c|}
\hline
\rowcolor[HTML]{C0C0C0} 
\textbf{MAD Algorithms}    & \textbf{Data Medium} & \textbf{BPCER @ MACER = 10\%} \\ \hline
                           & Digital              & 90.15                         \\ \cline{2-3} 
                           & PS-1                 & 75.98                         \\ \cline{2-3} 
\multirow{-3}{*}{ResNet50 \cite{Venkatesh-DeepColorMAD-IPTA-2019}} & PS-2                & 35.78                         \\ \hline
                           & Digital              & 52.48                         \\ \cline{2-3} 
                           & PS-1                 & 54.54                         \\ \cline{2-3} 
\multirow{-3}{*}{VGG-19 \cite{Raghavendra-DNNMorphingDetection-CVPR-2017}}   & PS-2                & \textbf{30.56}                         \\ \hline
                           & Digital              & 95.64                         \\ \cline{2-3} 
                           & PS-1                 & 84.45                         \\ \cline{2-3} 
\multirow{-3}{*}{ViT \cite{zhang2024generalized}}      & PS-2                & 75.78                         \\ \hline \hline
                           & Digital              & \textbf{29.87}                         \\ \cline{2-3} 
                           & PS-1                 & \textbf{43.15}                         \\ \cline{2-3} 
\multirow{-3}{*}{CLIP (Proposed Framework)}     & PS-2                & {90.87}                         \\ \hline \hline
\end{tabular}
}
\caption{Qualitative evaluation of the proposed explainable S-MAD with SOTS evaluated with zero-shot experiments.}
    \label{Table1} 
\end{table}
\begin{figure*}[]
  \centering
  \includegraphics[width=0.8\linewidth]{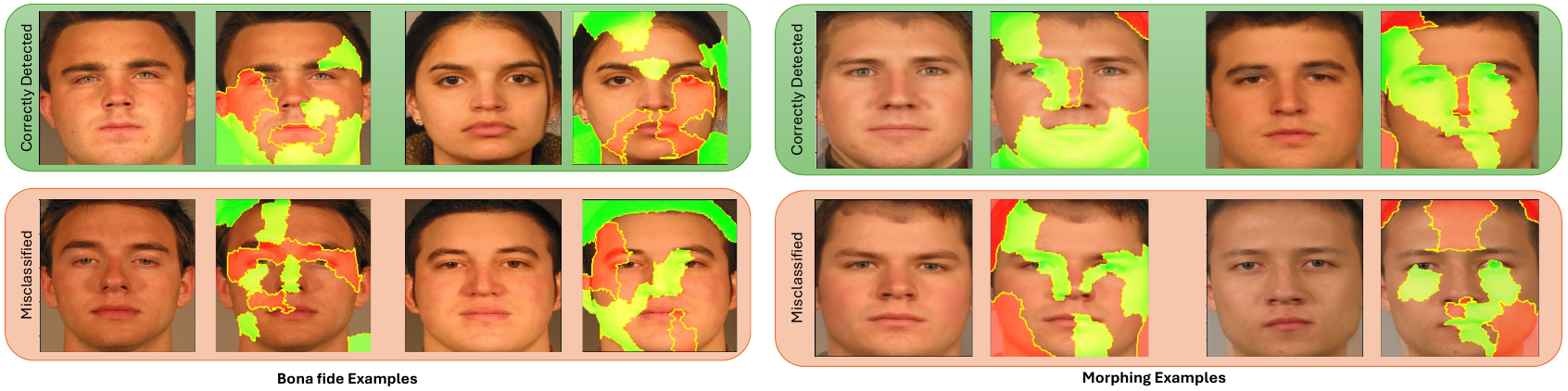}
  \caption{Lime explainability on CLIP video encoder on both bona fide and morphing
examples. Left figure indicates the LIME results for bona fide samples with correct
and not correct detection. Right figure shows the correct and non-correct detection for
morphing samples.}
  \label{fig:XLIME}
\end{figure*}

Table \ref{Table1} shows the quantitative performance of the proposed and SOTA neural networks with zero-shot evaluation. For simplicity, we have included the results corresponding to prompt \#5, as it is the best-performing prompt within the proposed framework. 
The obtained results indicate that (a) the zero-shot evaluation of CNN-based pretrained networks (ResNet50 and VGG-19) indicates good performance in detecting morphing attacks constructed using PS-2. (b) The transformer methods did not indicate good performance in detecting the morphing attacks across all the three mediums. (c) The proposed framework indicates the best results with digital and PS-1  mediums. Thus, indicating the best performance in the zero shot evaluation.

Figure \ref{fig:XLIME} shows the LIME-based interpretability results of the video encoder based on the ViT employed in the CLIP model. Qualitative results are presented for both bona fide and morphing examples with both correct and incorrect classifications.  As shown in Figure \ref{fig:XLIME},  (a) the correct classification of bona fide images can be attributed to the features selected in the whole images, while the incorrect classification is due to the extensive use of hair features; (b) the correct classification is mainly due to the importance of the feature in the eye region of the face, while the incorrect classification is due to the non-face features together with the scattered features of the face region.

Thus, based on the interpretability results, it is challenging to determine which features have contributed to the overall decision of the face morphing attack detector. However, the proposed multimodal approach can also predict the text snippet, which may provide additional information regarding the correct decision.  

\section{Conclusion}
\label{sec:conc}
In this paper, we demonstrated that image-language foundation models employed with zero-shot experiments can show improved generalization performance for morphing attack detection. We have shown that prompt engineering is important to improve  generalizable detection accuracy. We have analysed ten different prompts, including short and long versions, and benchmarked the morphing detection performance. Extensive experiments were performed on the morphing dataset constructed using five different morphing generation methods and three different data mediums (digital, high-quality print-scan, and low-quality print-scan) in zero-shot settings. Furthermore, we benchmarked the morphing attack detection performance of the image-language models with the image-only models. The obtained results indicate the best performance of the image-language model, particularly with short prompts. In the future, we plan to explore the fine-tuning of the vision model for morphing attack detection and to integrate the best vision model for morphing to boost generalizability. 

\section{Acknowledgment}
This work was supported by the European Commission [grant 101120657 "European Lighthouse to Manifest Trustworthy and Green AI" - ENFIELD]

\bibliographystyle{splncs04}
\bibliography{Face_Morph_references}

\begin{thebibliography}{10}
\providecommand{\url}[1]{\texttt{#1}}
\providecommand{\urlprefix}{URL }
\providecommand{\doi}[1]{https://doi.org/#1}

\bibitem{DNP-Printer}
{DNP} printer. \url{http://dnpphoto.com/en-us/Products/Printers/DS820A} (2020), accessed: October 2020

\bibitem{MoDiff}
Damer, N., Fang, M., Siebke, P., Kolf, J.N., Huber, M., Boutros, F.: Mordiff: Recognition vulnerability and attack detectability of face morphing attacks created by diffusion autoencoders. In: 2023 11th International Workshop on Biometrics and Forensics (IWBF). pp.~1--6 (2023). \doi{10.1109/IWBF57495.2023.10157869}

\bibitem{vit}
Dosovitskiy, A., Beyer, L., Kolesnikov, A., Weissenborn, D., Zhai, X., Unterthiner, T., Dehghani, M., Minderer, M., Heigold, G., Gelly, S., Uszkoreit, J., Houlsby, N.: An image is worth 16x16 words: Transformers for image recognition at scale. ArXiv  \textbf{abs/2010.11929} (2020), \url{https://api.semanticscholar.org/CorpusID:225039882}

\bibitem{Ferrara-TextureBlendingAndShapeWarpingInFaceMorphing-IEEE-BIOSIG-2019}
Ferrara, M., Franco, A., Maltoni, D.: Decoupling texture blending and shape warping in face morphing. In: Intl. Conf. of the Biometrics Special Interest Group ({BIOSIG}). IEEE (September 2019)

\bibitem{godage2022analyzing}
Godage, S.R., Løvåsdal, F., Venkatesh, S., Raja, K., Ramachandra, R., Busch, C.: Analyzing human observer ability in morphing attack detection—where do we stand? IEEE Transactions on Technology and Society  \textbf{4}(2),  125--145 (2023). \doi{10.1109/TTS.2022.3231450}

\bibitem{resnet50}
He, K., Zhang, X., Ren, S., Sun, J.: Deep residual learning for image recognition. In: Proceedings of the IEEE conference on computer vision and pattern recognition. pp. 770--778 (2016)

\bibitem{Landmark-face-morph}
landmark based~face morphing, F.: {Open CV}. \url{https://www.learnopencv.com/face-morph-using-opencv-cpp-python/}

\bibitem{Nist-Frvt-Morph}
NIST: {FRVT} morph web site. \url{https://pages.nist.gov/frvt/html/frvt_morph.html}

\bibitem{ICAO-9303-p1-2015}
Organization, I.C.A.: Machine readable passports -- part 1 -- introduction. \url{http://www.icao.int/publications/Documents/9303_p1_cons_en.pdf} (2015)

\bibitem{Phillips-OverviewFaceRecognitionGrandChallengeFRGC-CVPR-2005}
{Phillips}, P.J., {Flynn}, P.J., {Scruggs}, T., {Bowyer}, K.W., {Jin Chang}, {Hoffman}, K., {Marques}, J., {Jaesik Min}, {Worek}, W.: Overview of the face recognition grand challenge. In: 2005 IEEE Computer Society Conference on Computer Vision and Pattern Recognition (CVPR'05). pp. 947--954 vol. 1 (June 2005). \doi{10.1109/CVPR.2005.268}

\bibitem{clip}
Radford, A., Kim, J.W., Hallacy, C., Ramesh, A., Goh, G., Agarwal, S., Sastry, G., Askell, A., Mishkin, P., Clark, J., Krueger, G., Sutskever, I.: Learning transferable visual models from natural language supervision. In: Meila, M., Zhang, T. (eds.) Proceedings of the 38th International Conference on Machine Learning. Proceedings of Machine Learning Research, vol.~139, pp. 8748--8763. PMLR (18--24 Jul 2021), \url{https://proceedings.mlr.press/v139/radford21a.html}

\bibitem{radford2021learning}
Radford, A., Kim, J.W., Hallacy, C., Ramesh, A., Goh, G., Agarwal, S., Sastry, G., Askell, A., Mishkin, P., Clark, J., et~al.: Learning transferable visual models from natural language supervision. In: International conference on machine learning. pp. 8748--8763. PMLR (2021)

\bibitem{Raghavendra-DetectingMorphedFace-BTAS-2016}
Raghavendra, R., Raja, K., Busch, C.: Detecting morphed face images. In: 2016 {IEEE} 8th Intl. Conf. on Biometrics: Theory, Applications and Systems ({BTAS}). 8th {IEEE} Intl. Conf. on Biometrics: Theory, Applications and Systems (BTAS-2016), IEEE (September 2016)

\bibitem{Raghavendra-DNNMorphingDetection-CVPR-2017}
Raghavendra, R., Raja, K., Venkatesh, S., Busch, C.: Transferable deep-{CNN} features for detecting digital and print-scanned morphed face images. In: {IEEE} Conf. on Computer Vision and Pattern Recognition Workshops ({CVPRW}). pp. 1822--1830 (2017)

\bibitem{Raghavendra-DetectingFaceMorphing-CVIP-2018}
Raghavendra, R., Venkatesh, S., Raja, K., Busch, C.: Detecting face morphing attacks with collaborative representation of steerable features. In: Intl. Conf. on Computer Vision and Image Processing {(CVIP)} (September 2018)

\bibitem{RAJA2022104535}
Raja, K., Gupta, G., Venkatesh, S., Ramachandra, R., Busch, C.: Towards generalized morphing attack detection by learning residuals. Image and Vision Computing  \textbf{126},  104535 (2022). \doi{https://doi.org/10.1016/j.imavis.2022.104535}, \url{https://www.sciencedirect.com/science/article/pii/S0262885622001640}

\bibitem{ramachandra2024multispectral}
Ramachandra, R., Venkatesh, S., Damer, N., Vetrekar, N.: Multispectral imaging for differential face morphing attack detection: A preliminary study. In: Proceedings of the IEEE/CVF Winter Conference on Applications of Computer Vision. pp. 6185--6193 (2024)

\bibitem{10157461}
Ramachandra, R., Venkatesh, S., Jaswal, G., Li, G.: Vulnerability of face morphing attacks: A case study on lookalike and identical twins. In: 2023 11th International Workshop on Biometrics and Forensics (IWBF). pp.~1--6 (2023). \doi{10.1109/IWBF57495.2023.10157461}

\bibitem{10162056}
Ramachandra, R., Venkatesh, S., Li, G., Raja, K.: Differential newborn face morphing attack detection using wavelet scatter network. In: 2023 5th International Conference on Bio-engineering for Smart Technologies (BioSMART). pp.~1--4 (2023). \doi{10.1109/BioSMART58455.2023.10162056}

\bibitem{computers10090117}
Seibold, C., Hilsmann, A., Eisert, P.: Feature focus: Towards explainable and transparent deep face morphing attack detectors. Computers  \textbf{10}(9) (2021). \doi{10.3390/computers10090117}, \url{https://www.mdpi.com/2073-431X/10/9/117}

\bibitem{GradCam}
Selvaraju, R.R., Cogswell, M., Das, A., Vedantam, R., Parikh, D., Batra, D.: Grad-cam: Visual explanations from deep networks via gradient-based localization. In: Proceedings of the IEEE international conference on computer vision. pp. 618--626 (2017)

\bibitem{singh20233d}
Singh, J.M., Ramachandra, R.: 3d face morphing attacks: Generation, vulnerability and detection. IEEE Transactions on Biometrics, Behavior, and Identity Science  (2023)

\bibitem{10224168}
Singh, J.M., Venkatesh, S., Ramachandra, R.: Robust face morphing attack detection using fusion of multiple features and classification techniques. In: 2023 26th International Conference on Information Fusion (FUSION). pp.~1--8 (2023). \doi{10.23919/FUSION52260.2023.10224168}

\bibitem{srivatsan2023flip}
Srivatsan, K., Naseer, M., Nandakumar, K.: Flip: Cross-domain face anti-spoofing with language guidance. In: Proceedings of the IEEE/CVF International Conference on Computer Vision. pp. 19685--19696 (2023)

\bibitem{10157534}
Tapia, J.E., Busch, C.: Face feature visualisation of single morphing attack detection. In: 2023 11th International Workshop on Biometrics and Forensics (IWBF). pp.~1--6 (2023). \doi{10.1109/IWBF57495.2023.10157534}

\bibitem{transformer_text}
Vaswani, A., Shazeer, N., Parmar, N., Uszkoreit, J., Jones, L., Gomez, A.N., Kaiser, L., Polosukhin, I.: Attention is all you need. In: Proceedings of the 31st International Conference on Neural Information Processing Systems. p. 6000–6010. NIPS'17, Curran Associates Inc., Red Hook, NY, USA (2017)

\bibitem{Venkatesh-MADSurvey-IEEETTS-2021}
Venkatesh, S., Raghavendra, R., Raja, K., Busch, C.: Face morphing attack generation and detection: A comprehensive survey. IEEE Transactions on Technology and Society  \textbf{2}(3),  128--145 (March 2021). \doi{10.1109/TTS.2021.3066254}

\bibitem{Venkatesh-DeepColorMAD-IPTA-2019}
Venkatesh, S., Raghavendra, R., Raja, K., Spreeuwers, L., Veldhuis, R., Busch, C.: Morphed face detection based on deep color residual noise. In: 9th Intl. Conf. on Image Processing Theory, Tools and Applications ({IPTA}). IEEE (November 2019)

\bibitem{zhang-MIPGAN-TBIOM-2021}
Zhang, H., Venkatesh, S., Raghavendra, R., Raja, K., Damer, N., Busch, C.: {MIPGAN}—{G}enerating strong and high quality morphing attacks using identity prior driven {GAN}. IEEE Transactions on Biometrics, Behavior, and Identity Science  \textbf{3}(3),  365--383 (2021). \doi{10.1109/TBIOM.2021.3072349}

\bibitem{zhang2024generalized}
Zhang, H., Ramachandra, R., Raja, K., Busch, C.: Generalized single-image-based morphing attack detection using deep representations from vision transformer. In: Proceedings of the IEEE/CVF Conference on Computer Vision and Pattern Recognition. pp. 1510--1518 (2024)

\bibitem{PIPE}
Zhang, Haoyu;~Ramachandra, R.R.K.B.C.: Morph-pipe: Plugging in identity prior to enhance face morphing attack based on diffusion model. In: Norsk IKT-konferanse for forskning og utdanning (NISK). vol.~3, pp.~1--6 (2023)

\end{thebibliography}
\end{document}